\documentclass[11pt]{article}

\usepackage[final]{acl}

\usepackage{times}
\usepackage{latexsym}

\usepackage[T1]{fontenc}
\usepackage[utf8]{inputenc}

\usepackage{microtype}

\usepackage{inconsolata}

\usepackage{graphicx}

\usepackage{multirow}

\usepackage{booktabs}

\usepackage{amsmath}

\usepackage{float}

\usepackage{url}

\title{Archaeology at SemEval-2026 Task 13:\\Fine-Tuning Pre-Trained Code Models\\for AI-Generated Code Detection}

\author{Jany-Gabriel Ispas \and Sergiu Nisioi \\
  Human Language Technologies Research Center \\
  Faculty of Mathematics and Computer Science \\
  University of Bucharest \\
  \texttt{iani.ispas@gmail.com, sergiu.nisioi@unibuc.ro}}

\begin{document}
\maketitle
\begin{abstract}
This paper describes the system submitted by team \textbf{Archaeology} to SemEval-2026 Task~13 on AI-generated code detection.
The shared task consists of three subtasks; we participate in Subtask-A (binary classification: human-written vs.\ AI-generated code) and Subtask-B (11-class attribution of the generating model).
Starting from a TF-IDF and Logistic Regression baseline, we fine-tune four pre-trained code models (CodeBERT, GraphCodeBERT, UniXcoder, and CodeT5+) with separate strategies for each subtask.
For Subtask-A, we use leave-one-language-out cross-validation, code augmentation, chunked inference with trimmed-mean aggregation, and threshold calibration on a difficult dataset.
For Subtask-B, we use sandwich token packing, class-balanced loss, and multi-seed ensembling with test-time augmentation.
Our best submissions obtain macro-F1 scores of 0.737 on Subtask-A (6th/81 teams) and 0.422 on Subtask-B (7th/34 teams).
\end{abstract}

\section{Introduction}

Large language models are increasingly used to generate source code, which makes it important to detect whether a given code snippet is written by a human or produced by a machine.
Reliable detection has applications in academic integrity, software supply-chain security, and intellectual-property attribution.

SemEval-2026 Task~13 \cite{orel-etal-2026-semeval-2026} addresses this problem with three subtasks.
We participate in the first two: Subtask-A is a binary classification task where the goal is to predict whether a code snippet is human-written or AI-generated, and Subtask-B is an 11-class attribution task where the goal is to identify which model produced a given AI-generated code snippet.
Both subtasks use macro-averaged F1 as the evaluation metric.

We first train a TF-IDF and Logistic Regression \cite{cox1958regression,scikit-learn} baseline, then fine-tune four pre-trained code models: CodeBERT \cite{feng2020codebertpretrainedmodelprogramming}, GraphCodeBERT \cite{guo2021graphcodebertpretrainingcoderepresentations}, UniXcoder \cite{guo2022unixcoderunifiedcrossmodalpretraining}, and CodeT5+ \cite{wang2023codet5opencodelarge}.
The first three share the RoBERTa-base architecture (125M parameters), while CodeT5+ is an encoder-decoder model with 220M parameters.
All four are pre-trained on large-scale code corpora.
We design separate strategies for each subtask, motivated by their different challenges.
Subtask-A requires generalization to out-of-domain (OOD) programming languages not seen during training, since the training data contains only Python, C++, and Java while the test set includes several additional languages.
Subtask-B requires handling long code sequences and an extremely imbalanced class distribution (225:1 ratio between the largest and smallest class).

Our main contributions are:
(1) a chunked inference pipeline with trimmed-mean aggregation for binary code classification;
(2) a threshold calibration strategy that blends out-of-fold thresholds with thresholds from a difficult dataset built via cross-validated Logistic Regression; and
(3) a sandwich token packing approach with test-time augmentation for multi-class code attribution. Our code is publicly available.\footnote{\url{https://github.com/iani-ispas/Archaeology-SemEval2026}}

\section{Data and Tasks}

%We briefly describe the task setup and the key properties of the data that motivate our approach. For full details on the shared task, we refer the reader to \cite{orel-etal-2026-semeval-2026}.

\subsection{Subtask-A: Binary Classification}

\begin{figure}[H]
\centering
\includegraphics[width=\columnwidth]{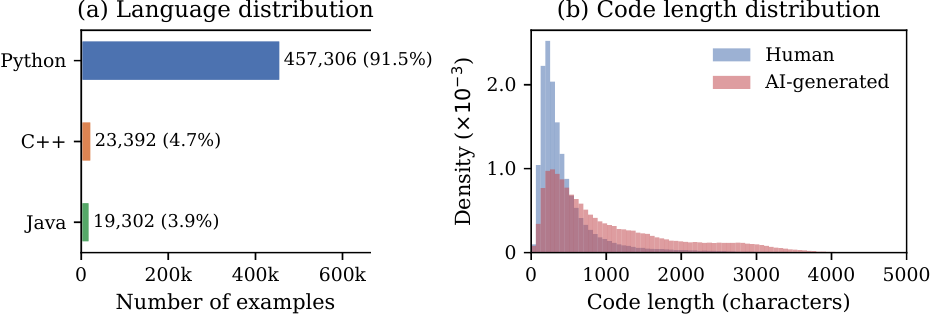}
\caption{Subtask-A training data. (a) Language distribution shows that Python dominates with 91.5\%. (b) Code length distributions differ between human-written and AI-generated code: AI code is longer and more uniform.}
\label{fig:subtask_a_data}
\end{figure}

The training set contains 500k code snippets labeled as human-written (0) or AI-generated (1), with an additional 100k validation examples.
The labels are roughly balanced (47.7\% human, 52.3\% AI).
However, the language distribution is heavily skewed: Python accounts for 91.5\% of the training data, while C++ (4.7\%) and Java (3.9\%) together make up less than 9\% (Figure~\ref{fig:subtask_a_data}a).
Within each language, the label split is approximately 50/50.

An important property of the data is that AI-generated code tends to be longer and more uniform in length than human-written code (Figure~\ref{fig:subtask_a_data}b).
AI-generated code has a mean length of 1,053 characters (median 726) compared to 600 characters (median 319) for human code.
Human code also has a much higher maximum length (475k vs. 12k characters), reflecting greater variability.
However, this pattern reverses for C++ and Java, where human-written code is longer on average than AI-generated code (Table~\ref{tab:code_length_a}).

The test set contains 500k examples and includes code in OOD languages not present in the training data, such as JavaScript, Go, PHP, C\#, and C.
The test set also has longer code on average (mean 1,421 characters vs. 837 in training), consistent with the presence of different, more verbose languages.

\subsection{Subtask-B: Multi-Class Attribution}

\begin{figure}[H]
\centering
\includegraphics[width=\columnwidth]{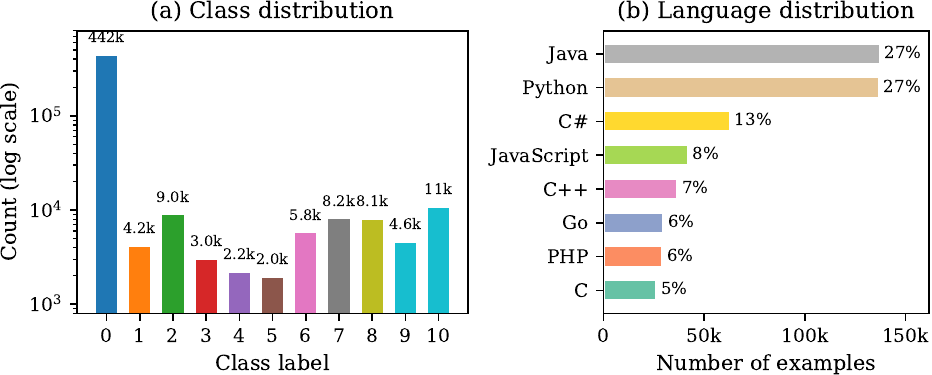}
\caption{Subtask-B training data. (a) Class distribution on a log scale shows the extreme imbalance: class~0 has 442k examples while the smallest class has under 2k. (b) Unlike Subtask-A, the 8 training languages are more evenly distributed.}
\label{fig:subtask_b_data}
\end{figure}

The training set contains 500k code snippets labeled with one of 11 generator model identifiers, plus 100k validation examples.
The 11 classes correspond to: (i) fully human-written code (class~0); and (ii--xi) code generated by one of ten LLM families: DeepSeek-AI \cite{deepseekai2024deepseekcoderv2breakingbarrierclosedsource}, Qwen \cite{hui2024qwen25codertechnicalreport}, 01-ai (Yi) \cite{ai2025yiopenfoundationmodels}, BigCode (StarCoder) \cite{li2023starcodersourceyou}, Gemma \cite{gemmateam2024gemmaopenmodelsbased}, Phi \cite{abdin2024phi3technicalreporthighly}, Meta-LLaMA \cite{touvron2023llamaopenefficientfoundation}, IBM-Granite \cite{mishra2024granitecodemodelsfamily}, Mistral \cite{jiang2023mistral7b}, and OpenAI (GPT) \cite{openai2024gpt4technicalreport}.
The class distribution is extremely imbalanced (Figure~\ref{fig:subtask_b_data}a): class~0 accounts for 88.4\% of the data (442k examples), while the smallest class (class~5) has only 1,968 examples, giving an imbalance ratio of approximately 225:1.

Unlike Subtask-A, the Subtask-B training data covers 8 programming languages (Python, Java, C\#, JavaScript, C++, Go, PHP, C), with a more balanced distribution (Figure~\ref{fig:subtask_b_data}b).
Java and Python each account for roughly 27\%, while the remaining languages contribute between 5\% and 13\%.

\begin{figure}[H]
\centering
\includegraphics[width=\columnwidth]{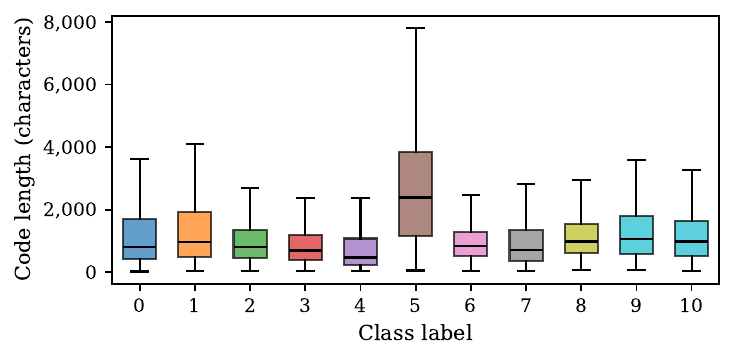}
\caption{Code length distribution per class in Subtask-B training data. Class~5 produces much longer code than all other classes.}
\label{fig:subtask_b_codelength}
\end{figure}

Code length varies across generator labels (Figure~\ref{fig:subtask_b_codelength}): most classes produce code with mean lengths between 800 and 1,500 characters, but class~5 has a mean of 2,694 characters, which is much longer than the rest.
The test set has a mean code length of 1,538 characters.

\section{Method}

\subsection{Baseline: TF-IDF + Logistic Regression}

As a baseline, we train a Logistic Regression classifier on TF-IDF features extracted from the raw code text.
We use character n-grams (bigrams through 5-grams) within word boundaries, retaining up to 100k features with sublinear TF weighting.
Code is truncated to the first 10,000 characters before feature extraction.
For Subtask-B, we use balanced class weights to account for the extreme label imbalance.
This baseline captures surface-level lexical patterns but does not use any pre-trained code representations.

\subsection{Pre-Trained Code Models}

We use three encoder-only models from the CodeBERT family, all sharing the RoBERTa-base architecture (12 transformer layers, 125M parameters), plus one encoder-decoder model:

\begin{itemize}
\item \textbf{CodeBERT} \cite{feng2020codebertpretrainedmodelprogramming}: pre-trained with masked language modeling (MLM) and replaced token detection on six programming languages.
\item \textbf{GraphCodeBERT} \cite{guo2021graphcodebertpretrainingcoderepresentations}: extends CodeBERT by incorporating data-flow graph structure during pre-training, which gives the model awareness of variable dependencies.
\item \textbf{UniXcoder} \cite{guo2022unixcoderunifiedcrossmodalpretraining}: a unified model that supports both understanding and generation tasks, pre-trained with cross-modal objectives including code-comment and AST-based tasks.
\item \textbf{CodeT5+} \cite{wang2023codet5opencodelarge}: an encoder-decoder model (220M parameters) from the CodeT5 family, pre-trained with a mixture of denoising and causal language modeling objectives on both unimodal code data and bimodal code-text data.
\end{itemize}

Each model is fine-tuned independently with a linear classification head on top of the \texttt{[CLS]} token representation (or the encoder output for CodeT5+).
All models use label smoothing of 0.02 for Subtask-A training.
Full hyperparameter settings are listed in Appendix~\ref{app:hparams}.

\subsection{Subtask-A: Binary Classification}

\subsubsection{Leave-One-Language-Out Cross-Validation}

The training data for Subtask-A contains only three languages (Python, C++, Java), but the test set includes several additional ones.
To simulate this OOD setting during training, we use leave-one-language-out (LOLO) cross-validation.
Each of the three folds trains on two languages and holds out the third for out-of-fold (OOF) prediction.
This produces three models per seed, each of which has never seen one of the training languages, providing a signal for how the model behaves on unseen languages.

Since Python dominates the training data (91.5\%), training on raw data would heavily bias the model toward Python-specific patterns.
To address this, we balance the training set within each fold by sampling equal numbers of examples per language and per label.
For folds that include Python, we sample up to 80k examples per language; for folds without Python, we sample up to 40k per language.
Underrepresented language-label pairs are oversampled with replacement.

\subsubsection{Code Augmentation}

To improve robustness to surface-level variation across languages, we apply three stochastic augmentations during training:

\begin{enumerate}
\item \textbf{Comment stripping} ($p{=}0.10$): removes block comments (\texttt{/* */}), line comments (\texttt{//}), and full-line hash comments (\texttt{\#}).
\item \textbf{Literal masking} ($p{=}0.15$): replaces string and numeric literals with placeholder tokens (\texttt{<STR>}, \texttt{<NUM>}).
\item \textbf{Header dropping} ($p{=}0.10$): removes the leading block of license headers, import statements, and docstrings (up to 40 lines).
\end{enumerate}

These augmentations reduce the model's reliance on language-specific surface features (such as comment syntax or import conventions) that would not transfer to unseen languages.

We also apply deterministic normalization to all data: removing markdown code fences (a source of training data leakage), normalizing tabs to spaces, trimming trailing whitespace, and compressing excessive blank lines.
For long code, we sample a random character-level chunk during training with a bias toward the beginning (50\%) or a random position (30\%) or the end (20\%).

\subsubsection{Chunked Inference}

At inference time, code snippets that exceed the chunk size are split into overlapping chunks.
For the three RoBERTa-based models, the chunk size is 900 characters with 120 characters of overlap; for CodeT5+, it is 1,024 characters with 150 characters of overlap.
The chunks always include the head and tail of the code, with uniformly spaced middle chunks filling the remaining budget (up to 6 chunks per example).
Each chunk is tokenized independently and classified by the model.
The per-chunk softmax probability for the AI-generated class is aggregated using a 50/50 blend of two statistics: a trimmed mean (dropping the minimum and maximum values when there are 3 or more chunks) and a top-2 mean (the average of the two highest chunk probabilities).
This combination is robust against noisy chunks while still giving weight to the most confident predictions.

This design choice is motivated by the observation that the test set contains much longer code than the training set (mean 1,421 vs. 837 characters).
A fixed truncation would miss most of the code, so chunking allows the model to consider the full snippet.

\subsubsection{Threshold Calibration}

The binary decision threshold is calibrated in two stages.

\textbf{Stage 1: OOF threshold.}
After LOLO cross-validation, we collect the predicted probabilities on all held-out language folds.
We then search over 99 quantile-based candidate thresholds to find the one that maximizes macro-F1 on these OOF predictions.
This threshold reflects the model's behavior on unseen languages.

\textbf{Stage 2: Difficult dataset threshold.}
We construct a small difficult dataset of 1,000 examples by running 3-fold cross-validation with Logistic Regression on the full training data and sampling from the misclassified examples.
These are examples where a simple linear model fails, and they serve as a proxy for difficult test examples.
We compute a second threshold on this dataset by searching for the macro-F1-maximizing value.

The final threshold is an equal-weight blend: $\tau = 0.5 \cdot \tau_\text{OOF} + 0.5 \cdot \tau_\text{diff}$.
We also compute per-language thresholds on the difficult dataset using a rule-based language detector and apply these at inference time.
For test examples whose language is not recognized, we fall back to the global blended threshold.
This per-language calibration is important because, as shown in Table~\ref{tab:code_length_a}, the characteristics of human vs. AI code differ across languages.

\subsection{Subtask-B: Multi-Class Classification}

\subsubsection{Sandwich Token Packing}

Source code in Subtask-B is often longer than what the model can process in a single forward pass (median 801 characters for class~0, up to 2,393 for class~5).
Rather than truncating from the beginning or end, we use a ``sandwich'' packing strategy: the input sequence consists of the first $h$ tokens (head), a delimiter token sequence (\texttt{/*<MID\_SNIP>*/}), and the last $t$ tokens (tail), where $h + t$ equals the available token budget after accounting for special tokens.
This preserves information from both the beginning and end of the code, which is useful because code structure and style patterns can appear anywhere.

During training, the head fraction $h/(h+t)$ is randomly sampled from $\{0.50, 0.60, 0.70\}$ for each example, which acts as a data augmentation.
During evaluation, it is fixed at 0.60.
Before tokenization, code is also capped at 24,000 characters for the RoBERTa-based models (10,000 for CodeT5+), keeping both head and tail, to speed up encoding.

For UniXcoder, we prepend the special \texttt{<encoder-only>} mode token, following the model's recommended usage for classification tasks.

\subsubsection{Class-Balanced Loss}

The extreme class imbalance (225:1 ratio, see Figure~\ref{fig:subtask_b_data}a) means that standard cross-entropy loss would be dominated by the majority class.
We use class-balanced (CB) loss weights based on the effective number of samples \cite{cui2019classbalancedlossbasedeffective}:
\begin{equation}
w_c = \frac{1 - \beta}{1 - \beta^{n_c}}
\end{equation}
where $n_c$ is the number of training examples for class $c$ and $\beta = 0.9995$.
The weights are normalized to have unit mean.
This gives higher weight to underrepresented classes while avoiding the extreme upweighting that inverse-frequency weights would produce.

\subsubsection{Multi-Seed Ensemble and Test-Time Augmentation}

All three RoBERTa-based models apply test-time augmentation (TTA): at inference, each model is run three times with different head fractions (0.50, 0.60, 0.70) and the predicted softmax distributions are averaged.
For UniXcoder (our best Subtask-B model), we additionally train three models with different random seeds (42, 43, 44), giving $3 \times 3 = 9$ forward passes per example.
For CodeBERT and GraphCodeBERT, we use a single seed with TTA (3 forward passes each).

CodeT5+ uses a single seed and a fixed head fraction of 0.60 (no TTA), and trains for only 1 epoch (vs. 3 for the other models) due to its larger size.

All Subtask-B models use a cosine learning rate scheduler with early stopping (patience of 1 epoch), so training may terminate before the configured number of epochs if validation macro-F1 does not improve.

\section{Results and Analysis}

\subsection{Main Results}

Table~\ref{tab:results} shows the macro-F1 scores for each model on both subtasks.
The TF-IDF + LR baseline achieves 0.266 on Subtask-A and 0.255 on Subtask-B, showing that surface-level features alone are not enough for either task.
All four fine-tuned models clearly outperform this baseline.
Our best Subtask-A submission (CodeBERT) ranks \textbf{6th out of 81 teams}, and our best Subtask-B submission (UniXcoder) ranks \textbf{7th out of 34 teams}.

\begin{table}[H]
\centering
\resizebox{\columnwidth}{!}{
\begin{tabular}{lcc}
\toprule
\textbf{Model} & \textbf{Subtask-A} & \textbf{Subtask-B} \\
\midrule
TF-IDF + LR (baseline) & 0.266 & 0.255 \\
\midrule
CodeBERT        & \textbf{0.737} & 0.392 \\
GraphCodeBERT   & 0.730          & 0.391 \\
UniXcoder       & 0.669          & \textbf{0.422} \\
CodeT5+         & 0.727          & 0.385          \\
\bottomrule
\end{tabular}
}
\caption{Macro-F1 on the test set. Bold indicates the best result per subtask.}
\label{tab:results}
\end{table}

\subsection{Subtask-A Analysis}

CodeBERT and GraphCodeBERT perform similarly (0.737 and 0.730), CodeT5+ is close behind (0.727), while UniXcoder lags at 0.669.
The three RoBERTa-based models share the same architecture and training pipeline, so their differences come from pre-training.
CodeBERT and GraphCodeBERT are pre-trained with objectives that focus on code understanding (MLM and data-flow prediction), which aligns well with the binary detection task.
UniXcoder is pre-trained with additional generation objectives, which may not help for a task that requires only encoding.
CodeT5+, despite being a larger encoder-decoder model (220M parameters) with a higher learning rate ($3 \times 10^{-5}$ vs. $2 \times 10^{-5}$) and longer chunk size (1,024 vs. 900 characters), does not outperform the smaller encoder-only models, which suggests that model size and input length alone do not explain performance on this task.

All models use short maximum sequence lengths (128 tokens for the RoBERTa-based models, 256 for CodeT5+), relying on the chunked inference pipeline to handle long code.
The trimmed mean reduces the influence of noisy chunks, while the top-2 mean ensures that if any chunk strongly indicates AI-generated code, that signal is preserved.

The threshold calibration is important because the test data may have a different label distribution from the training set.
The difficult dataset (misclassified examples from Logistic Regression cross-validation) provides a more representative calibration target than the raw training data.
The per-language thresholds further help because different languages have different characteristics: for instance, Python AI code tends to be longer than human code, while in C++ and Java the opposite is true (Table~\ref{tab:code_length_a}), which affects the optimal decision boundary.

\subsection{Subtask-B Analysis}

UniXcoder achieves the highest macro-F1 (0.422), followed by CodeBERT (0.392), GraphCodeBERT (0.391), CodeT5+ (0.385).
The ranking differs from Subtask-A, which is interesting given that the three encoder-only models share the same base architecture.

Several factors explain UniXcoder's advantage on Subtask-B.
First, we use a maximum sequence length of 1,024 tokens for UniXcoder (and CodeT5+), compared to 512 for CodeBERT and GraphCodeBERT.
This allows the sandwich packing to retain more of the code, which matters because distinguishing between AI generators likely depends on subtle stylistic differences distributed across the full code snippet.
Second, UniXcoder uses a 3-seed ensemble with TTA (9 forward passes per example), while CodeBERT and GraphCodeBERT use a single seed with TTA (3 passes each), and CodeT5+ uses a single pass.
Third, UniXcoder's \texttt{<encoder-only>} mode token may help by signaling to the model that the task is classification rather than generation.

The class-balanced loss is important for this subtask due to the extreme class imbalance.
Without it, the model tends to predict the majority class (class~0, which is 88.4\% of the training data) for most examples, resulting in poor macro-F1.
Since macro-F1 weights all classes equally regardless of size, the reweighting ensures the model learns to distinguish the smaller classes as well.

\subsection{Error Analysis}

We analyze errors on a labeled subset of the test data.
For Subtask-A, the models make more errors on OOD languages than on the three training languages.
For example, CodeBERT's error rate on JavaScript and C is 20\% and 19\% respectively, compared to 14\% on C++ and Java.
The gap is larger for other models: CodeT5+ reaches a 38\% error rate on JavaScript.
The per-language threshold calibration helps reduce this gap, but the models remain less reliable on languages not seen during training.

For Subtask-B, class~0 (human-written code) and class~10 are reliably classified (F1 $\geq$ 0.80 across all models), but most AI-generator classes have much lower F1.
Class~1 is the hardest to classify (F1 $\leq$ 0.11 for all models), despite having more training data (4,162 examples) than classes~3, 4, or~5.
This indicates that class size alone does not determine difficulty; rather, the key challenge is distinguishing generators that produce stylistically similar code.
The most frequent confusions involve classes~2, 6, 7, 8, and~10, which are consistently misclassified as each other across all four models.

\section{Conclusions and Future Work}

We present a system for AI-generated code detection that fine-tunes pre-trained code models with task-specific strategies.
For binary classification, leave-one-language-out cross-validation and threshold calibration on a difficult dataset help generalize to unseen programming languages.
For multi-class attribution, sandwich token packing with class-balanced loss and multi-seed ensembling address the challenges of long code and class imbalance.

Our results show that the choice of pre-training objective matters more than model size: CodeBERT and GraphCodeBERT, with their focus on code understanding, perform best on binary detection, while UniXcoder, combined with a longer context window and ensembling, performs best on multi-class attribution.
CodeT5+, despite having 220M parameters (vs. 125M for the others), does not achieve the top score on either subtask.

Future work includes exploring larger pre-trained models (e.g., CodeLlama), incorporating syntactic features such as abstract syntax trees, and investigating contrastive learning objectives for better generator discrimination.

% ------------------------------------------------------------------
\section*{Limitations}

Our approach uses only pre-trained models with 125M to 220M parameters and maximum sequence lengths of 128 to 1,024 tokens.
Larger models with longer contexts may capture additional signals.
The chunked inference strategy for Subtask-A introduces a trade-off between coverage and noise: more chunks cover more of the code but may dilute the signal from the most discriminative regions.
Our threshold calibration assumes access to a representative difficult dataset at submission time, which may not always be available.
Finally, we do not perform extensive ablation experiments due to computational constraints; the contribution of each individual component (augmentations, chunk aggregation, per-language thresholds) remains to be quantified in isolation.

\section*{Acknowledgments}
This research is supported by InstRead: Research Instruments for the Text Complexity, Simplification and Readability Assessment  CNCS - UEFISCDI project number PN-IV-P2-2.1-TE-2023-2007.

\bibliography{custom}

\appendix

\section{Dataset Details}
\label{app:data}

\begin{table}[H]
\centering
\small
\begin{tabular}{llrr}
\toprule
\textbf{Language} & \textbf{Label} & \textbf{Count} & \textbf{Mean len.} \\
\midrule
\multirow{2}{*}{Python} & Human & 218,103 & 442 \\
                         & AI    & 239,203 & 1,027 \\
\midrule
\multirow{2}{*}{C++}    & Human & 11,147  & 1,943 \\
                         & AI    & 12,245  & 1,254 \\
\midrule
\multirow{2}{*}{Java}   & Human & 9,225   & 2,716 \\
                         & AI    & 10,077  & 1,409 \\
\bottomrule
\end{tabular}
\caption{Subtask-A training set: example counts and mean code length (in characters) by language and label. Python AI code is longer than human code, but this pattern reverses for C++ and Java.}
\label{tab:code_length_a}
\end{table}

Table~\ref{tab:code_length_a} shows the code length statistics for each language-label combination in the Subtask-A training set.
The reversal of the length pattern between Python and the other two languages is one motivation for using per-language thresholds during calibration.

Table~\ref{tab:class_dist_b} shows the full class distribution for Subtask-B, including the count and percentage for each of the 11 classes.

\begin{table}[H]
\centering
\small
\begin{tabular}{crrc}
\toprule
\textbf{Class} & \textbf{Count} & \textbf{\%} & \textbf{Mean len.} \\
\midrule
0  & 442,096 & 88.42 & 1,443 \\
1  & 4,162   & 0.83  & 1,404 \\
2  & 8,993   & 1.80  & 986 \\
3  & 3,029   & 0.61  & 898 \\
4  & 2,227   & 0.45  & 842 \\
5  & 1,968   & 0.39  & 2,694 \\
6  & 5,783   & 1.16  & 989 \\
7  & 8,197   & 1.64  & 1,018 \\
8  & 8,127   & 1.63  & 1,197 \\
9  & 4,608   & 0.92  & 1,304 \\
10 & 10,810  & 2.16  & 1,156 \\
\bottomrule
\end{tabular}
\caption{Subtask-B training set: class distribution and mean code length per class. Class~0 dominates with 88.4\% of examples. Class~5 produces the longest code.}
\label{tab:class_dist_b}
\end{table}

\section{Hyperparameters}
\label{app:hparams}

Tables~\ref{tab:hparams_a} and~\ref{tab:hparams_b} list the full hyperparameter settings for each model on both subtasks.

\begin{table}[H]
\centering
\small
\begin{tabular}{lcc}
\toprule
\textbf{Hyperparameter} & \textbf{RoBERTa models} & \textbf{CodeT5+} \\
\midrule
Max sequence length   & 128     & 256 \\
Batch size            & 32      & 16 \\
Gradient accumulation & 1       & 2 \\
Learning rate         & $2 \times 10^{-5}$ & $3 \times 10^{-5}$ \\
Epochs                & 1       & 1 \\
Warmup ratio          & 0.06    & 0.06 \\
Weight decay          & 0.01    & 0.01 \\
Label smoothing       & 0.02    & 0.02 \\
Chunk size (chars)    & 900     & 1,024 \\
Overlap (chars)       & 120     & 150 \\
Max chunks per code   & 6       & 6 \\
Seeds                 & \{1337\}& \{1337\} \\
\bottomrule
\end{tabular}
\caption{Subtask-A hyperparameters. All three RoBERTa-based models (CodeBERT, GraphCodeBERT, UniXcoder) share the same settings.}
\label{tab:hparams_a}
\end{table}

\begin{table}[H]
\centering
\small
\resizebox{\columnwidth}{!}{
\begin{tabular}{lcccc}
\toprule
\textbf{Hyperparameter} & \textbf{CodeBERT} & \textbf{GraphCB} & \textbf{UniXcoder} & \textbf{CodeT5+} \\
\midrule
Max seq.\ length    & 512  & 512  & 1,024 & 1,024 \\
Batch size          & 8    & 8    & 4     & 16 \\
Grad.\ accum.       & 4    & 4    & 8     & 2 \\
Learning rate       & 2e-5 & 2e-5 & 1.5e-5 & 1.5e-5 \\
Epochs              & 3    & 3    & 3     & 1 \\
CB $\beta$          & \multicolumn{4}{c}{0.9995} \\
Early stop patience & \multicolumn{4}{c}{1 epoch} \\
LR scheduler        & \multicolumn{4}{c}{cosine} \\
Pre-token.\ cap      & 24k  & 24k  & 24k   & 10k \\
Seeds               & \{42\} & \{42\} & \{42,43,44\} & \{42\} \\
TTA fractions       & 3    & 3    & 3     & 1 \\
\bottomrule
\end{tabular}
}
\caption{Subtask-B hyperparameters. ``GraphCB'' abbreviates GraphCodeBERT. TTA fractions refers to the number of head fractions (0.50, 0.60, 0.70) used at test time; CodeT5+ uses only 0.60.}
\label{tab:hparams_b}
\end{table}

\end{document}